\documentclass[letterpaper, 10 pt, conference]{ieeeconf}  % Comment this line out if you need a4paper

\IEEEoverridecommandlockouts                              % This command is only needed if 
\usepackage{graphicx} % for pdf, bitmapped graphics files
\usepackage{amssymb}  % assumes amsmath package installed
\usepackage{amsmath} % assumes amsmath package installed

\title{\LARGE \bf
Three-Dimensional Anatomical Data Generation Based on Artificial Neural Networks*}
\author{Ann-Sophia Müller$^{1,2}$, Moonkwang Jeong$^{1}$, Meng Zhang$^{1}$, Jiyuan Tian$^{1}$, \\ Arkadiusz Miernik$^{3}$, Stefanie Speidel$^{4}$, and Tian Qiu$^{1,5,*}$% <-this % stops a space
\thanks{*This work was partially funded by the German Cancer Research Center (DKFZ) and the Cyber Valley Research Fund (CyVy-RF-2020-09).}% <-this % stops a space
\thanks{$^{1}$A-S. Müller, M. Jeong, M. Zhang, J. Tian, and T. Qiu are with the Division of Smart Technologies for Tumor Therapy, 
        German Cancer Research Center (DKFZ) Site Dresden, Blasewitzer Str. 80, 01307 Dresden, Germany}%
\thanks{$^{2}$A-S. Müller is also with the Faculty of Computer Science, Dresden University of Technology, 01187 Dresden, Germany}%
\thanks{$^{3}$A. Miernik is with the Department of Urology, Faculty of Medicine, University of Freiburg - Medical Centre, Hugstetterstraße 55, 79106, Freiburg}
\thanks{$^{4}$S. Speidel is with the Division of Translational Surgical Oncology, National Center for Tumor Diseases (NCT/UCC) Dresden, 01307 Dresden, Germany}
\thanks{$^{5}$T. Qiu is also with the Faculty of Medicine Carl Gustav Carus, Dresden University of Technology, 01307 Dresden, Germany, and the Faculty of Electrical and Computer Engineering, Dresden University of Technology, 01187 Dresden, Germany
        {\tt\small tian.qiu@dkfz.de}}%
}

\begin{document}

\maketitle
\thispagestyle{empty}
\pagestyle{empty}

%%%%%%%%%%%%%%%%%%%%%%%%%%%%%%%%%%%%%%%%%%%%%%%%%%%%%%%%%%%%%%%%%%%%%%%%%%%%%%%%
\begin{abstract}

Surgical planning and training based on machine learning requires a large amount of 3D anatomical models reconstructed from medical imaging, which is currently one of the major bottlenecks. Obtaining these data from real patients and during surgery is very demanding, if even possible, due to legal, ethical, and technical challenges. It is especially difficult for soft tissue organs with poor imaging contrast, such as the prostate. To overcome these challenges, we present a novel workflow for automated 3D anatomical data generation using data obtained from physical organ models. We additionally use a 3D Generative Adversarial Network (GAN) to obtain a manifold of 3D models useful for other downstream machine learning tasks that rely on 3D data. We demonstrate our workflow using an artificial prostate model made of biomimetic hydrogels with imaging contrast in multiple zones. This is used to physically simulate endoscopic surgery. For evaluation and 3D data generation, we place it into a customized ultrasound scanner that records the prostate before and after the procedure. A neural network is trained to segment the recorded ultrasound images, which outperforms conventional, non-learning-based computer vision techniques in terms of intersection over union (IoU). Based on the segmentations, a 3D mesh model is reconstructed, and performance feedback is provided.

\end{abstract}

%%%%%%%%%%%%%%%%%%%%%%%%%%%%%%%%%%%%%%%%%%%%%%%%%%%%%%%%%%%%%%%%%%%%%%%%%%%%%%%%
\section{INTRODUCTION}
\begin{figure*}[thp] 
\includegraphics[width=\textwidth]{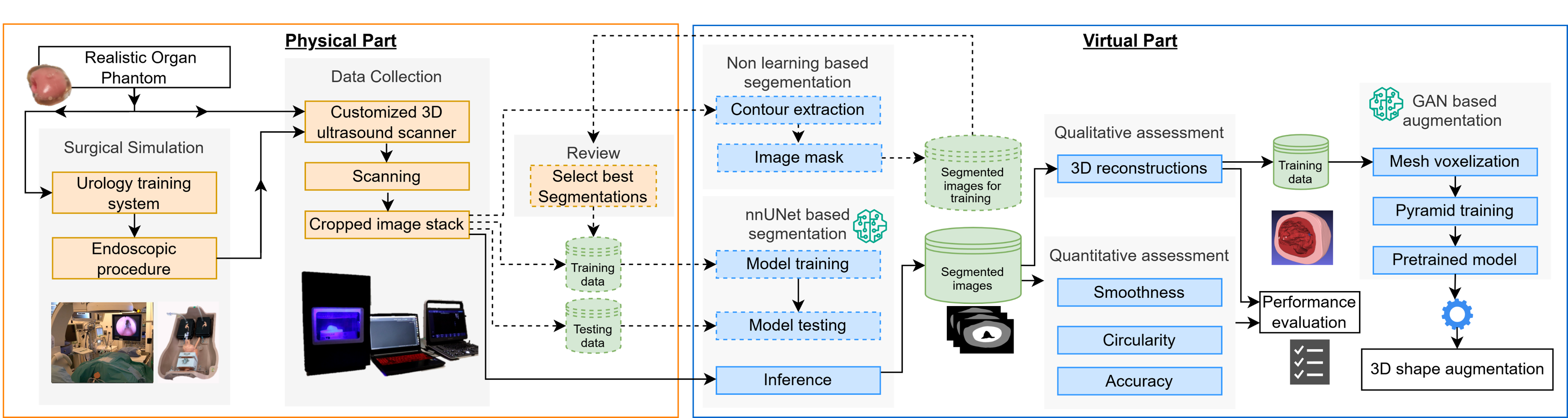}\caption{Overview of the workflow. The left side shows the components for physical simulation and data collection. The right side shows the virtual tasks for data segmentation, reconstruction, evaluation, and augmentation. Solid arrows represent the final workflow. Dashed arrows show the preceding steps necessary to build the final pipeline.} \label{fig1}
\end{figure*}
Learning by doing is fruitful, especially for surgical training. However, unlike in most professions, a mistake during surgery can have severe consequences for the patient, and there is often just a fine line between success and permanent damage. Virtual Reality (VR) or Augmented Reality (AR) surgical simulations provide a low-risk training opportunity for surgeons to practice their skills. However, these solutions are limited by the availability of detailed 3D anatomical data, which is essential for creating realistic 3D visual representations of the anatomy. Unfortunately, these data are sparse due to several factors: 1) structured medical images, especially during surgery, are difficult to collect due to legal, ethical, and technical challenges. 2) Processing and annotation of imaging data require laborious manual intervention by trained medical experts. This is especially the case for soft tissue organs, as they are less perceptible in common imaging modalities such as MRI or ultrasound. 3) Augmenting the 3D shapes to obtain more variation, again, requires domain-specific expertise and hands-on intervention knowledge.

VR and AR surgical simulations often lack realistic haptic feedback, limiting the sense of touch and interaction with virtual tissues. To provide a realistic surgical simulation, we developed cyber-physical phantoms of soft tissue organs like the liver~\cite{tan2021soft}, the bladder~\cite{choi2021soft}, and the prostate~\cite{choi2020high}. These phantoms are fabricated using hydrogels and offer realistic biomimetic mechanical properties and behavior when interacting with surgical instruments. The phantoms are placed in a urinary tract training system which altogether ensures high-fidelity haptic and visual feedback.

To overcome the data annotation difficulties and to obtain more enhanced visualization under ultrasound compared to real soft tissue organs, the phantoms are induced with an ultrasound contrast agent. This allows highlighting task-specific regions under ultrasound to provide insights that are hard to obtain from real soft tissue organs. Because there is no need for ethical or legal approval, this system can be exploited to quickly retrieve an extensive amount of data at any point during the surgical simulation. This makes our system highly compatible for data-driven imaging, such as machine learning-based segmentation~\cite{garcia2017review}.

We aim to provide 3D reconstructions of organ models and quantitative feedback based on 2D ultrasonic images obtained from the phantoms using a customized scanning setup. To offer a general workflow compatible with different setups (e.g., different ultrasound machines) and datasets, we rely on a versatile segmentation method. Isensee~\textit{et al.} provide such a method by introducing the nnUNet~\cite{isensee2021nnu}, a semantic segmentation model adaptable to various datasets, evaluated on 23 biomedical datasets. Furthermore, we demonstrate how our workflow can be extended to quickly obtain a manifold of organ models based on a single 3D reconstruction by using the 3D GAN proposed by Wu~\textit{et al.}~\cite{wu2022learning}. This eliminates the need for additional data collection or manual intervention when augmented versions of a sample are required. These abundant anatomical models are also useful for other downstream biomedical machine learning tasks that require a large amount of 3D anatomical data~\cite{schneider2021medmeshcnn} or the rapid rendering of varying surgical scenes for VR surgical simulation~\cite{harders2008surgical}, and surgical robotic vision and navigation~\cite{li2020anatomical}~\cite{hein2024creating}.

For the first time, we present a comprehensive end-to-end workflow for surgical training in Transurethral Resection of the Prostate (TURP). Our approach integrates automated data collection using a customized ultrasound scanner with data processing for 3D reconstruction and augmentation. We successfully tested this workflow with two different ultrasound machines, highlighting the generalizability of our method. Our contributions are 1) realistic surgical simulations offered by our phantoms with subsequent automated and high-quality data collection using a customized ultrasound scanning machine. 2) The integration of two artificial neural networks (ANNs), an nnUNet~\cite{isensee2021nnu} to segment, and a 3D GAN~\cite{wu2022learning} to augment these data for 3D visualizations. 3) The automated workflow is furthermore useful for surgical training with VR and AR, but also may benefit the training of surgical robots in the future.

\section{METHODOLOGY AND MATERIALS }
The workflow comprises physical and virtual components cf. Fig.~\ref{fig1}. Surgical simulations are performed on organ phantoms, and data is collected using a customized ultrasound scanning setup. The collected images are then processed in the virtual part for automated assessment, 3D mesh reconstruction, and augmentation.
\subsection{Physical Part}
\subsubsection{Surgical Simulation} 
\begin{figure*}[ht] 
\centering
\includegraphics[width=\textwidth]{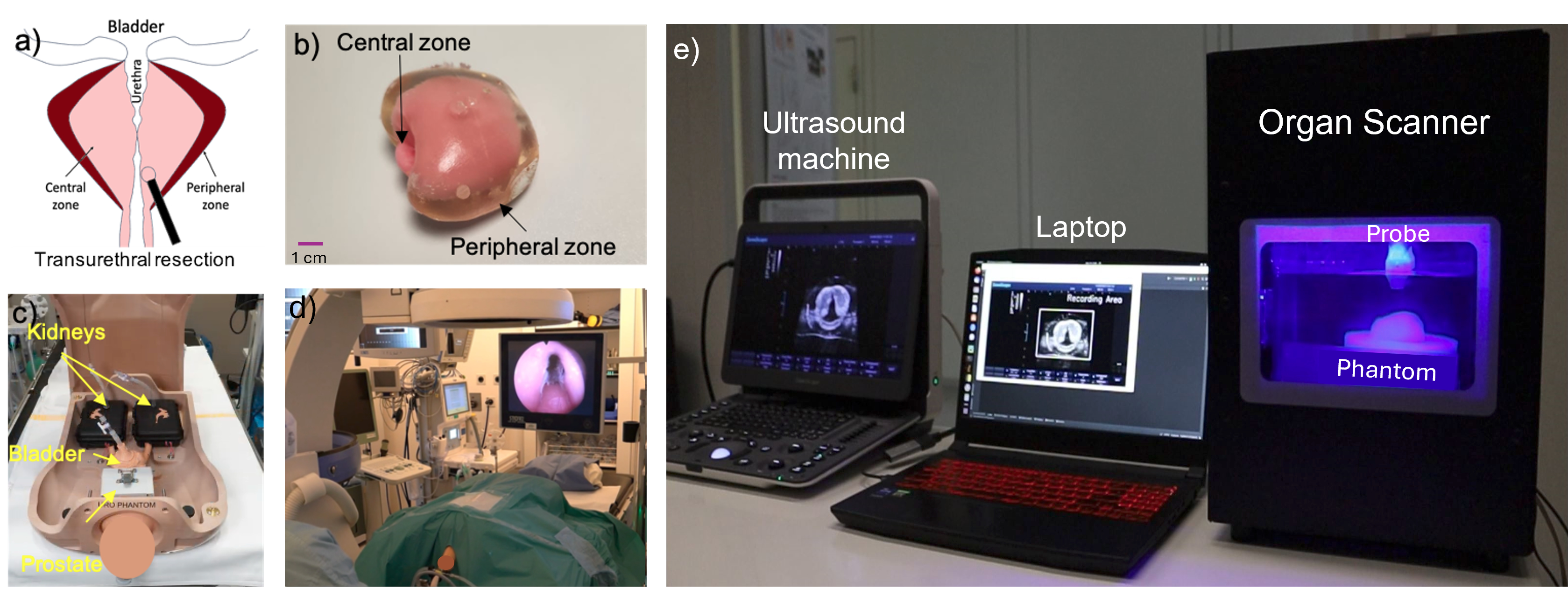}\caption{Overview of the physical setup: a) schematic of a prostate with BPH. b) The two-zone prostate phantom made of hydrogels. c) Endo Urology Trainer. d) Surgical simulation of TURP on the phantom. e) The complete system for ultrasound scanning and data generation, including our Organ Scanner to move the ultrasound imaging probe along the longitudinal axis for a 3D scan of the prostate phantom.} \label{fig2}
\end{figure*}

We build upon our previous work~\cite{choi2020high}, on physically simulating the TURP procedure. This is the current gold standard to treat Benign Prostatic Hyperplasia (BPH), a condition in which the urethra is blocked due to an enlarged central zone of the prostate. The endoscopic surgery to resect the prostate by electrocauterization is illustrated in Fig.~\ref{fig2}a. To achieve optimal results, the central zone should be removed as much as possible without perforating the surrounding peripheral zone, as this would result in irreversible damage to the patient. Mastering this skill requires practice, which we provide using the realistic model of the prostate phantom encompassing two zones, cf. Fig.~\ref{fig2}b. This model is inserted into a dedicated Endo Urology Trainer (QUANTITATIVE SURGICAL GmbH, Mannheim, Germany) cf. Fig.~\ref{fig2}c. Surgical instruments (SHARK, RICHARD WOLF GmbH, Knittlingen, Germany) that are used in actual clinical procedures are used to simulate the TURP procedure and enucleation on the prostate models, cf. Fig.~\ref{fig2}d.
\subsubsection{Imaging system and automated data collection} We set up an imaging system cf. Fig.~\ref{fig2}e consisting of three parts: a) a portable SonoScape E1 ultrasound machine with a linear probe L746 (both SONOSCAPE MEDICAL CORP, China), b) our Organ Scanner, to move the ultrasound probe along the longitudinal axis of the prostate to obtain a stack of images, and c) a consumer-grade MSI Katana GF66 12UGS-261 laptop (MSI, Taiwan) with an RTX 3070 Ti laptop-GPU (NVIDIA, USA) and Core i7-12700H processor (INTEL, USA) to capture, store and process the images. The Organ Scanner consists of a linear stage (LMD17S13RF15-120, STEPPERONLINE, China) for the translational motion and a stepper motor (14HS11-1004D-PG19-AR3, STEPPERONLINE, China) for the rotational motion of the ultrasound probe. A customized 3D-printed gripper is attached to the rotational motor to hold the probe. A glass container filled with water is placed beneath the linear stage, and the phantom is placed under water at a fixed position. The water tank can easily be slid in or out of the scanner to allow for quick change of samples.

At any time during the procedure or afterwards, the prostate phantom can be retrieved from the urology training system and placed into the scanner. A button press triggers the scanning motion, which is defined to be 60 mm in the longitudinal direction of the prostate phantom. The sample is placed such that the scan starts at the upper urethral opening, cf. Fig. \ref{fig2}a and e. An algorithm running on the consumer-grade laptop captures the video output of the ultrasound machine at 30 FPS in HD resolution. Capturing is implemented in Python (version 3.11) using OpenCV~\cite{opencvlibrary} (version 4.9.0) and using an HDMI capturing device (Video capture Card HDMI to USB, PAPEASO SHENZHEN YIJIASAN TECHNOLOGY Co., Ltd, China).

The algorithm automatically detects the beginning of the scanning by monitoring changes in the average gray value of the capture input. Once this value exceeds a predefined threshold, the input stream is collected as image slices to form a dataset. This workflow allows for the quick retrieval of numerous datasets. We furthermore collected data from a second commercial ultrasound imaging machine (LOGIO P6, GE HEALTHCARE, USA), to show the general usability of our approach.
\begin{figure*}[ht] 
\centering
\includegraphics[width=\textwidth]{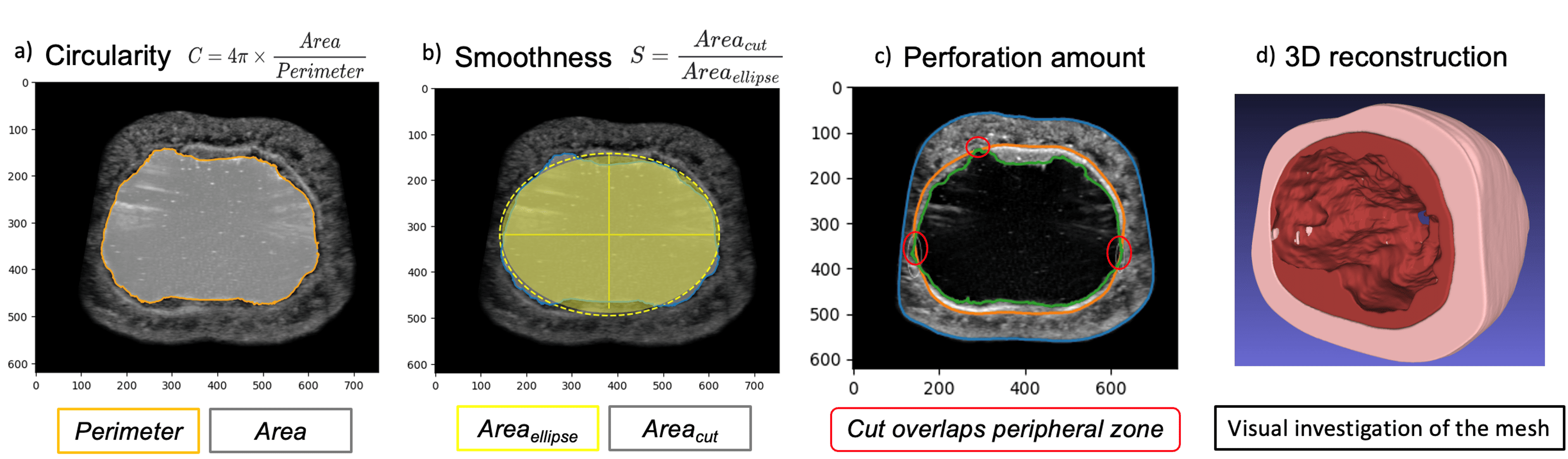}\caption{Automated workflow for 2D image performance assessment and 3D reconstruction. a) Evaluation of the circularity of the resection, b) Evaluation of the smoothness of the resection, c) Evaluation of the perforation amount, and d) 3D reconstruction to qualitatively assess the performance on a full 3D mesh model of the resected sample} \label{fig3}
\end{figure*}
\subsection{Virtual Part}
\subsubsection{Initial semi-automated image processing and annotation} 
The primary goal of the segmentation is to accurately delineate the peripheral and central zones of the prostate, with particular emphasis on the detailed reconstruction of the central zone, as its resection is the focus of the TURP procedures. To maintain a cost-efficient workflow, we rely on ultrasound as the imaging modality. However, achieving the detailed visualization of tissue structures, necessary for segmentation in real soft tissue organs under ultrasound, is challenging or even impossible. To address this, the prostate phantoms are engineered with imaging contrast agents in different zones~\cite{choi2020high}, making them clearly distinguishable under ultrasound. This facilitates not only the segmentation but also the preceding annotation procedure.

While there is a low cost in capturing a large amount of imaging data using our scanner, this data must be annotated before it can be used for data-driven segmentation in the final workflow. We incorporate a preceding semi-autonomous annotation stage into the workflow to alleviate this workload. Captured images are automatically segmented using traditional, non-learning-based computer vision techniques. First, the outer contour of the prostate is extracted (blue contour in Fig. \ref{fig3}c) by smoothing the image and using the active contour method as implemented in the scikit-image~\cite{van2014scikit} library (version 0.22.0), initialized with a rectangular contour that shrinks to fit the boundary. Since the position of the prostate phantom is fixed, the position and diameter of the initial contour can be defined in advance. In subsequent images, the previously extracted contour is used as the initial contour. The peripheral zone’s inner contour (orange contour in Fig. \ref{fig3}c) is then extracted similarly. Finally, the resection contour (green contour in Fig. \ref{fig3}c) is identified using contrast enhancement based on morphological operations~\cite{soille1999morphological}, morphological Chan-Vese segmentation~\cite{getreuer2012chan}, and a flood-fill operation. The extracted contours are used to form a mask showing the peripheral zone, central zone, and resection space and background. This mask can be used as ground truth label for training a data-driven segmentation algorithm. The annotation procedure is semi-automatic because, as a final step, a human needs to review the generated masks and select the most suitable ones.

\subsubsection{Automated ANN based segmentation} 
We aim to offer a general workflow that reliably reconstructs organ models across different datasets and imaging setups. Therefore, a robust and versatile segmentation method is needed. A requirement that can not be met using our non-learning-based methods. In recent years, researchers have shifted towards ANN segmentation methods in the biomedical field, replacing traditional computer vision algorithms. A robust and widely used semantic segmentation model in medical imaging that has the ability for self-configuration, making it compatible with various setups, was introduced by Isensee \textit{et al.} with the nnUNet~\cite{isensee2021nnu}. This flexibility is highly useful for our final workflow and thus an nnUNet is trained to replace the non-data-driven segmentation method.

To generate training data, we use the semi-automated image processing and annotation stage of the workflow. Phantoms are rapidly scanned before and after resection using our Organ Scanner with the SonoScape ultrasound machine. Captured images are cropped and automatically segmented using the non-learning-based methods resulting in two image stacks of 85 images (600px height, 950px width), for each scanned sample. Ten suitable pairs of capture data and non-learning-based segmentations are selected to serve as data to train and test the nnUNet~\cite{isensee2021nnu}. This dataset includes five resected and five unresected samples. To introduce greater variability, within both, the resected and unresected samples, half were recorded with a dynamic range of 48 dB, while the other half were recorded with a dynamic range of 120 dB.

We employed the nnUNet v2.2.1 using the 3D full-resolution U-Net configuration, trained with five-fold cross-validation. No manual modifications to hyperparameters or architectural components of the model were made. Training was performed on a workstation equipped with an RTX 4080 GPU (NVIDIA, USA) and S-1700 i9-13900K Processor (INTEL, USA). Because the final workflow requires only inference and post-processing, we deployed the pre-trained nnUNet~\cite{isensee2021nnu} on the same consumer-grade laptop used for scanning and data collection, enabling the entire workflow to run on a single device. The resulting image stacks were converted into 3D mesh models using meshlib (v2.4.2.198) for interactive visualization. This allows for qualitative evaluation of the procedures performed on the phantom by examining the mesh. In addition, a quantitative performance assessment is provided by scoring each frame according to metrics predefined by experienced surgeons (see Fig.~\ref{fig3}).

%Segmentation of each sample was completed in under one minute on this system. 
\subsubsection{Automated ANN based 3D shape augmentation}\label{sec:augm}
Goodfellow \textit{et al.}~\cite{goodfellow2020generative} introduced GANs, a machine learning framework of two neural networks, a generator and a discriminator, competing against each other. The generator produces data, and the discriminator learns to distinguish real from fake. Through this adversarial process, the generator improves in producing realistic outputs. GANs are widely used for image generation and have become more prominent in the biomedical field, e.g., to augment sparse or limited datasets \cite{chen2022generative}.
In this work, we explore the application of a GAN for augmenting 3D anatomical models. Wu \textit{et al.}~\cite{wu2022learning} proposed a 3D GAN framework that can generate variations of 3D meshes by training on a single example. This is realized by using a multi-scale architecture designed to capture both; global structures and local details. By training on a voxel-pyramid of the input, the model learns a representation that preserves the overall shape of the input but encourages small variations. Using this method, we can generate a variety of 3D meshes of the physical phantom.

However, due to memory constraints, fine-grained features get lost during voxelization, making the peripheral and central zones less distinguishable. To address this issue, we use the following approach: Given the dataset that should be augmented, we first generate two meshes from it. One where the central zone is filled (i.e., for each image slice, the inner empty space is flood-filled). The second mesh is created by reconstructing only the resection area, which typically appears as empty space in 3D meshes but is now converted into a filled volume, with its boundary corresponding to the resection (cf. Fig. \ref{fig3} c, green boundary). By performing a Boolean subtraction of this second mesh from the first, we would obtain the same 3D reconstruction as if we had directly reconstructed the segmented images. However, this approach allows us to train the GAN only on the resection volume to augment many diverse volumes from it. After subtracting these augmented resection volumes from the first mesh, we obtain multiple variations of the original data while ensuring that the boundaries remain intact throughout the meshes. 

We adhered to the instructions described by Wu \textit{et al.}~\cite{wu2022learning} for data pre-processing, training, and shape generation. The model was trained across six scales, with a final voxel resolution of $128^3$, representing the dimensions of the 3D grid used to voxelize the mesh. Although higher resolutions are possible with more powerful hardware, we chose to stay within the limits of the consumer-grade laptop for a unified, portable workflow.
\begin{figure*}[ht] 
\centering
\includegraphics[width=15cm]{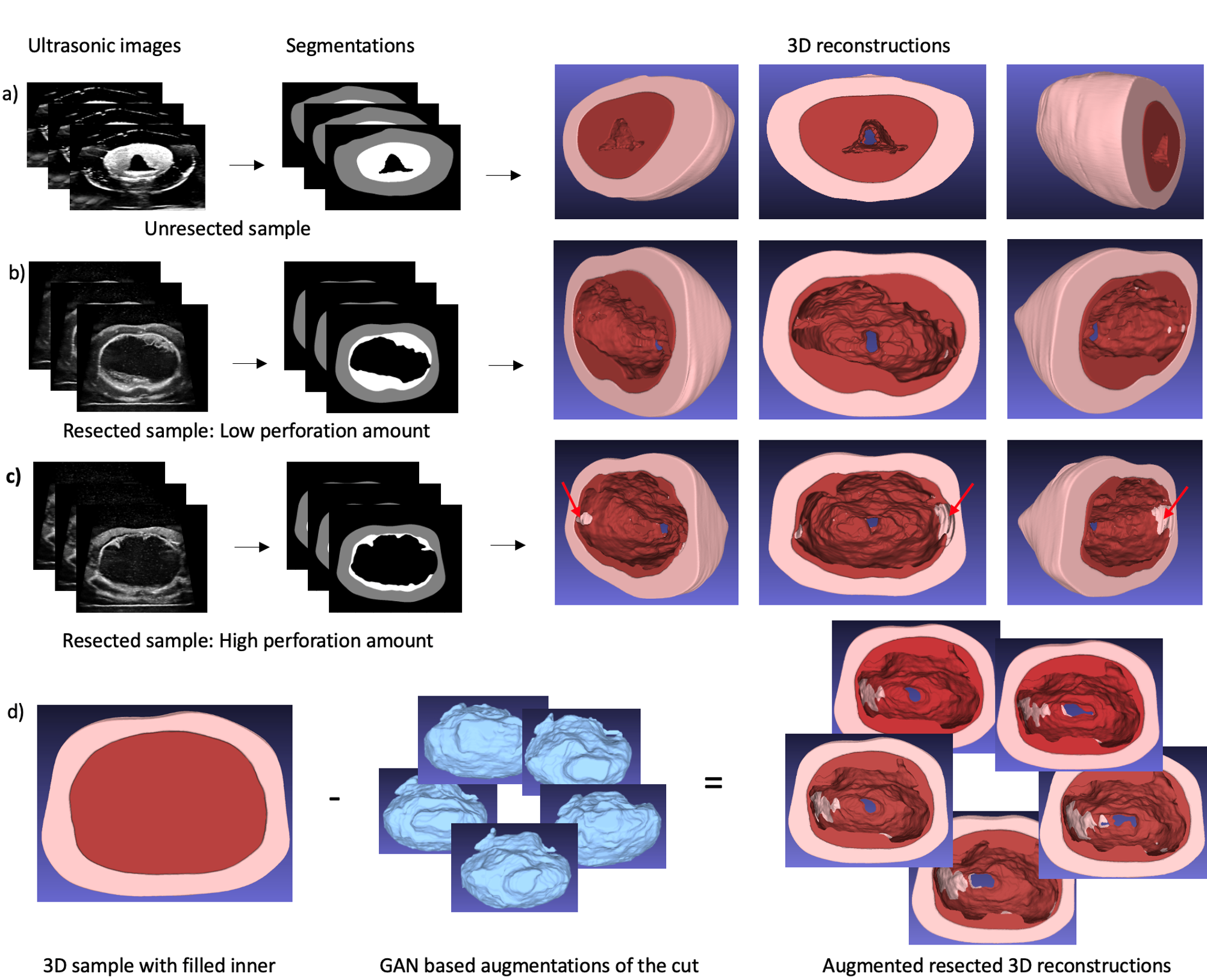}\caption{Examples of reconstructed and augmented multi-layer prostate 3D models. a) an unresected sample, b) a resected sample with little perforation, c) a resected sample with many perforations that are labeled by arrows in the 3D reconstruction, d) augmented anatomical shapes retrieved by subtracting different GAN-generated resection volumes from a constant peripheral and filled central zone volume} \label{fig4}
\end{figure*}

\section{Experimental Results}
\subsection{Evaluation of the segmentation methods}
To evaluate the quality of the semi-automatically generated annotations, we create additional non-learning-based segmentations and select five sets that we would consider as suitable ground truth labels. However, we do not include these sets for training the nnUNet~\cite{isensee2021nnu}. Instead, for the same captured data, we generate segmentations using the nnUNet. In addition, we manually segment the five selected sets to serve as a reference standard. We then compare the IoU scores of the nnUNet segmentations and the non-learning-based segmentations against the manually labeled reference. Because accurately segmenting the central zone is our primary objective, we provide a separate evaluation specifically for this region.
\begin{table}[h]
\caption{Comparison of the segmentation performance by nnUNet~\cite{isensee2021nnu} and best non-learning-based results. The numbers represent the mean IoU value $\pm$ standard deviation.}\label{tab1}
\centering
\setlength{\tabcolsep}{1em} % for the horizontal padding
{\renewcommand{\arraystretch}{1}% for the vertical padding
\begin{tabular}{llll} 
\hline\hline
Method       & Overall IoU~                                     & Central IoU                                      & Peripheral IoU                                    \\ 
\hline
nnUNet~\cite{isensee2021nnu}      & 0.81 $\pm$ 0.04 & 0.86 $\pm$ 0.05 & 0.76 $\pm$ 0.05  \\
non-learning & 0.78 $\pm$ 0.06 & 0.83 $\pm$ 0.05 & 0.74 $\pm$ 0.08  \\
\hline\hline
\end{tabular}
}
\end{table}

From table \ref{tab1}, we can draw two key observations. 1) The preselected non-learning-based segmentations achieve a high IoU score (i.e., 0.83 for the central zone), confirming their suitability as ground truth labels for training the nnUNet. This shows a key advantage of incorporating the imaging contrast agents into our phantoms: it significantly reduces the data annotation effort required for machine learning model training. 2) Despite being trained on a small dataset with automatically generated ground truth, not including these five datasets, the nnUNet outperforms the selected best non-learning-based segmentations, reflected in its higher IoU scores of 0.86 for the central zone. We expected the nnUNet to be limited by the quality of the generated ground truth, which is shown to be not perfect (IoU $<$ 1). However, this is not the case, which demonstrates that the neural network can effectively learn segmentation criteria from imperfect ground truth data. Additionally, we observed that the nnUNet successfully segmented datasets where non-learning-based methods had failed.

\subsection{Qualitative evaluation of the workflow}
We evaluated the quality and usefulness of our workflow by using it to assist a urologist in comparing three distinct surgical techniques for prostatic surgery. An experienced urologist performed each procedure on ten prostate phantoms. Subsequently, 30 samples were scanned using the Organ Scanner for quantitative analysis. Each frame was analyzed to extract three parameters: a) circularity of the resection area boundary, b) smoothness of the resection area surface, and c) perforation amount in the peripheral zone. These parameters are defined as illustrated in Fig.~\ref{fig3}a-c. A statistical analysis of these parameters, automatically extracted by our workflow, revealed the significant superiority of one surgical technique across all evaluated aspects. These findings align with the urologists' expectations, and provide quantitative feedback for surgical training and interactive 3D visualization (cf. Fig~\ref{fig3}d), showing the usefulness of our solution. 
%We evaluated the quality and usefulness of our workflow by using it to assist a urologist in comparing three distinct surgical techniques for prostatic surgery. An experienced urologist performed each procedure on ten prostate phantoms. Subsequently, 30 samples were scanned using the Organ Scanner for quantitative analysis. Each frame was analyzed to extract three parameters: a) circularity of the boundary of the resection area, b) smoothness of the surface of the resection area and c) amount of perforation in the peripheral zone. These parameters are defined as illustrated in Fig.~\ref{fig3}a-c. A statistical analysis of these parameters, automatically extracted by our workflow, revealed the significant superiority of one surgical technique in all the aspects evaluated. These findings align with the urologists' expectations, and provide quantitative feedback for surgical training and interactive 3D visualization (cf. Fig~\ref{fig3}d), showing the usefulness of our solution. 

The scans were collected using the GE ultrasound machine, resulting in image stacks of 130 slices (620×755px) which the nnUNet~\cite{isensee2021nnu} was successfully trained on, demonstrating the compatibility of our approach with different ultrasound systems. Applying the full workflow from scanning to 3D reconstruction using the pre-trained nnUNet enabled the retrieval of detailed 3D shapes accurately reflecting the anatomy of the scanned phantom and surgical performance, in under three minutes on a consumer-grade laptop. Further, we successfully augmented the resulting 3D reconstructions to generate meshes that closely resemble, yet remain distinct from the scanned prostate phantoms using the GAN proposed by Wu \textit{et al.}~\cite{wu2022learning}. Final results are depicted in Fig.~\ref{fig4}.
\section{DISCUSSION}
Given the frequent failure cases of traditional (non-learning-based) segmentation methods, where multiple attempts were needed to obtain just five usable segmentation sets, compared to nnUNet~\cite{isensee2021nnu}, which succeeded on all five datasets in a single run, we argue that nnUNet offers significantly more reliable and robust segmentations. Especially when considering that only the best traditional results are shown, yet the nnUNet still outperforms them. This makes it highly suitable for our fully automated, end-to-end workflow, supporting both performance assessment and 3D anatomical model generation. While conventional segmentation methods may produce inaccurate segmentation masks, our results show that with human guidance, they can still be useful for quickly generating high-quality annotations for learning-based methods. Although selecting the best segmentations requires some manual effort, this semi-autonomous stage is far less labor-intensive than manual slice-by-slice annotation. Moreover, once a single 3D reconstruction is completed, a large amount of anatomical data can be generated without repeated scanning. This may be useful for the rendering of varying surgical scenes or the training of surgical robots that need to quickly adapt to small but important changes in the environment.
\section{CONCLUSIONS}
We present a novel workflow for efficient collection, segmentation, and 3D reconstruction of anatomical data using surgical simulations on physical organ models. These segmentations and reconstructions provide quantitative and qualitative performance feedback to surgical trainees. Our findings show that ground truth annotations generated using non-learning-based computer vision methods can be used to train an nnUNet~\cite{isensee2021nnu} to perform the segmentation task in the final workflow. Our experiments further indicate, that the data-driven segmentation method results in more accurate and especially more robust segmentation results compared to the non-data-driven methods. Finally, the 3D reconstructions can be augmented using a GAN trained on a single sample to generate diverse anatomical structures. This variability is valuable for VR/AR surgical simulations (e.g., endoscopic navigation) and for training medical robots requiring large, diverse 3D datasets.
\addtolength{\textheight}{-12cm}   % This command serves to balance the column lengths

%%%%%%%%%%%%%%%%%%%%%%%%%%%%%%%%%%%%%%%%%%%%%%%%%%%%%%%%%%%%%%%%%%%%%%%%%%%%%%%%

%%%%%%%%%%%%%%%%%%%%%%%%%%%%%%%%%%%%%%%%%%%%%%%%%%%%%%%%%%%%%%%%%%%%%%%%%%%%%%%%

%%%%%%%%%%%%%%%%%%%%%%%%%%%%%%%%%%%%%%%%%%%%%%%%%%%%%%%%%%%%%%%%%%%%%%%%%%%%%%%%

\section*{ACKNOWLEDGMENT}

 The authors acknowledge laboratory supports by D. Li, and L. Boggaram-Naveen.

%%%%%%%%%%%%%%%%%%%%%%%%%%%%%%%%%%%%%%%%%%%%%%%%%%%%%%%%%%%%%%%%%%%%%%%%%%%%%%%%
\bibliographystyle{ieeetr}  % Or another style like IEEE, apa, etc.
\bibliography{IEEEfull}  % This is the .bib file without the .bib extension

\end{document}